# A REINFORCEMENT LEARNING HYPER-HEURISTIC IN MULTI-OBJECTIVE SINGLE POINT SEARCH WITH APPLICATION TO STRUCTURAL FAULT IDENTIFICATION

**Pei Cao**
Department of Mechanical Engineering
University of Connecticut
Storrs, CT 06269 USA
pei.cao@uconn.edu

**J. Tang**
Department of Mechanical Engineering
University of Connecticut
Storrs, CT 06269 USA
jiong.tang@uconn.edu

**ABSTRACT**

Multi-objective optimizations are frequently encountered in engineering practices. The solution techniques and parametric selections however are usually problem-specific. In this study we formulate a reinforcement learning hyper-heuristic scheme, and propose four low-level heuristics which can work coherently with the single point search algorithm MOSA/R (Multi-Objective Simulated Annealing Algorithm based on Re-seed) towards multi-objective optimization problems of general applications. Making use of the domination amount, crowding distance and hypervolume calculations, the proposed hyper-heuristic scheme can meet various optimization requirements adaptively and autonomously. The approach developed not only exhibits improved and more robust performance compared to AMOSA, NSGA-II and MOEA/D when applied to benchmark test cases, but also shows promising results when applied to a generic structural fault identification problem. The outcome of this research can be extended to a variety of design and manufacturing optimization applications.

**KEYWORDS:** Hyper-heuristic, multi-objective optimization, simulated annealing, structural fault identification.

## 1. INTRODUCTION

Many engineering optimization problems involve multiple types of goals, thus naturally present themselves as multi-objective problems. For example, the rapid advancement of sensing and measurement technologies has made it possible to realize structural fault identification in near real-time. Fault parameters in a structure are generally identified through matching measurements with model predictions in the parametric space. Since multiple measurements are usually involved, the identification can be cast into a multi-objective optimization problem.

Multi-objective optimization algorithms have been practically applied to a variety of applications, ranging from production scheduling (Wang et al, 2014; Lu et al, 2016), structural design (Kaveh and Laknejadi, 2013), performance improvement (Szollos et al, 2009), to structural fault pattern recognition (Cao et al, 2018a; 2018b) etc. The solution techniques, nevertheless, are often devised and evaluated for specific problem domains, which not only require in-depth understanding of the problem domain involved but are also difficult to be exercised to different instances. Even for the same type of problems, the formulation may need to be adjusted as more knowledge and insights are gained. The hyper-heuristic concept was therefore suggested (Cowling et al, 2000), aiming at producing general-purpose approaches. The terminology implies that a high-level scheme to select heuristic operators is incorporated as the detailed algorithms are being executed (Burke et al, 2009) given a particular problem and a number of low-level heuristics. Instead of finding good solutions, hyper-heuristic is more interested in adaptively finding good solution methods. Since its emergence, the subject has gained significant interests, and a number of studies of hyper-heuristic have been performed for multi-objective problems. Burke et al (2007) and Sabar et al (2011) proposed hyper-heuristic approaches to address timetabling and scheduling problems. Gomez and Terashima-Marin (2010), de Armas et al (2011) and Bai et al (2012) extended the hyper-heuristic method to handle packing and space allocating problems. Raad et al (2010) and McClymont and Keedwell (2011) used hyper-heuristics to water resource and distribution problems. Wang and Li (2010) and Vazquez-Rodrigues and Petrovic (2013) also applied hyper-heuristic framework to multi-objective benchmark problems such as DTLZ and WFG. More recently, Guizzo et al (2015) applied hyper-heuristic based multi-objective evolutionary algorithms to solve search-based software engineering problems. Hitomi and Selva (2015; 2016) investigated the effect of credit definition and aggregation strategies on multi-objective hyper-heuristics and used it to solve satellite optimization problems. Interested readers may refer to (Burke et al, 2013; Maashi et al, 2015) for more discussions about hyper-heuristic techniques and applications.

Typically, a hyper-heuristic framework involves: (1) a high-level selection strategy to iteratively select among low-level heuristics based on the performance; (2) a predefined repository of low-level heuristics; and (3) applying the heuristics selected into optimization and evaluating their performance. The selection mechanism in hyper-heuristics, which essentially ensures the objectivity, specifies the heuristic to apply in a given point of optimization without using any domain information. With this in mind, online learning hyper-heuristics usually take advantage of the concept of reinforcement learning for selection (Kaelbling et al, 1996;

Ozcan et al, 2012), as it aims to iteratively solve the heuristics selection task by weight adaptation through interactions with the search domain. The low-level heuristics correspond to a set of exploration rules, and each carries a utility value. The values are updated at each step based on the success of the chosen heuristic. An improving move is rewarded, while a worsening move is punished. The low-level heuristics can be embedded in single point search techniques, which are highly suited for these tasks because only one neighbor is analyzed for a choice decision (Nareyek, 2003). In a single point search-based hyper-heuristic framework, e.g., a simulated annealing (Kirkpatrick et al, 1983) based hyper-heuristic, an initial candidate solution goes through a set of successive stages repeatedly until termination.

The goal of this research is to develop a formulation for general-purpose multi-objective optimization framework. Specifically, we want to advance the state-of-the-art in Multi-Objective Simulated Annealing (MOSA) by incorporating hyper-heuristic systematically to improve both the generality and performance. We develop a reinforcement learning hyper-heuristic inspired by probability matching (Goldberg, 1990), which consists of a selection strategy and a credit assignment strategy. As discovered in previous investigations (Cao et al, 2016 and 2017), the solution quality/diversity as well as the robustness of the algorithm can be enhanced with re-seed schemes. The re-seed schemes, on the other hand, need to be tailored to fit specific problem formulation. Here in this research the re-seed schemes are treated as the low-level heuristics, empowering the algorithm to cover various scenarios. The performance and generality of the proposed approach are first demonstrated over benchmark testing cases DTLZ (Deb et al, 2002) and UF (Zhang et al, 2008) in comparison with popular multi-objective algorithms, namely, NSGA-II (Deb et al, 2002), AMOSA (Bandyopadhyay et al, 2008) and MOEA/D (Zhang and Li, 2007). The enhanced approach is then applied to structural fault identification, a highly promising application of MOSA, to examine the practical implementation. The main contributions of this paper are 1) to introduce a new reinforcement learning hyper-heuristic framework based on MOSA with re-seed; 2) to devise a new credit assignment strategy in high-level selection for heuristic performance evaluation; and 3) to provide insights on benchmark case studies and application.

## 2. ALGORITHMIC FOUNDATION
*2.1. Multi-Objective Optimization (MOO)*

Intuitively, Multi-Objective Optimization (MOO) could be facilitated by forming an alternative problem with a single, composite objective function using weighted sum. Single objective optimization techniques are then applied to this composite function to obtain a single optimal solution. However, the weighted sum methods have difficulties in selecting proper weight factors especially when there is no articulated *a priori* preference among objectives. Indeed, *a posteriori* preference articulation is usually preferred, because it allows a greater degree of separation between the optimization methodology and the decision-making process which also enables the algorithmic development process to be conducted independently of the application (Giagkiozos et al, 2015). Furthermore, instead of a single optimum produced by the weighted sum approach, MOO can yield a set of solutions exhibiting explicitly the tradeoff between different objectives.

The most well-known MOO methods are probably the Pareto-based ones that define optimality in a wider sense that no other solutions in the search space are superior to Pareto optimal solutions when all objectives are considered (Zitzler, 1999). A general MOO problem of $n$ objectives in the minimization sense is represented as:

$$\text{Minimize } \mathbf{y} = \boldsymbol{f}(\mathbf{x}) = (f_1(\mathbf{x}),...,f_n(\mathbf{x})) \quad (1)$$

where $\mathbf{x} = (x_1, x_2, ..., x_k) \in \mathbf{X}$ and $\mathbf{y} = (y_1, y_2, ..., y_n) \in \mathbf{Y}$. $\mathbf{x}$ is the decision vector of $k$ decision variables, and $\mathbf{y}$ is the objective vector. $\mathbf{X}$ denotes the decision space while $\mathbf{Y}$ is called the objective space. When two sets of decision vectors are compared, the concept of dominance is used. Assuming $\mathbf{a}$ and $\mathbf{b}$ are decision vectors, the concept of Pareto optimality can be defined as follows: $\mathbf{a}$ is said to dominate $\mathbf{b}$ if:

$$\forall i = \{1, 2, ..., n\} : f_i(\mathbf{a}) \leq f_i(\mathbf{b}) \quad (2)$$

and

$$\exists j = \{1, 2, ..., n\} : f_j(\mathbf{a}) < f_j(\mathbf{b}) \quad (3)$$

Refer to Table 1. Any objective function vector which is neither dominated by any other objective function vector of a set of Pareto-optimal solutions nor dominating any of them is called non-dominated with respect to that Pareto-optimal set (Goldberg, 1989; Zitzler, 1999). The solution that corresponds to the objective function vector is a member of Pareto-optimal set. Usually $\prec$ is used to denote domination relationship between two decision vectors (Table 1).

Table 1 Domination relations

| Relation | Symbol | Interpretation in objective space |
|---|---|---|
| **a** dominates **b** | $\mathbf{a} \prec \mathbf{b}$ | **a** is not worse than **b** in all objectives and better in at least one |
| **b** dominates **a** | $\mathbf{b} \prec \mathbf{a}$ | **b** is not worse than **a** in all objectives and better in at least one |
| Non-dominant to each other | $\mathbf{b} \cong \mathbf{a}$ | **a** is worse than **b** in some objectives but better in some other objectives |

*2.2. Multi-Objective Simulated Annealing (MOSA)*

Simulated annealing (Kirpatrick et al, 1983) is a heuristic technique drawing an analogy from physics annealing process. It was originally designed for solving single objective optimization problem, and then extended to multi-objective context. Engrand, who is among the very first to embed the concept of Pareto optimality with simulated annealing, proposed to maintain an external population archiving all non-dominated solutions during the solution procedure (Engrand, 1998). Several Multi-Objective Simulated Annealing (MOSA) algorithms that incorporate Pareto set (Nam and Park, 2000; Suman, 2004) have been developed. The acceptance criteria in these algorithms are all derived from the differential between new and current solutions. However, in the presence of Pareto set, merely comparing the new solution to the current solution appears to be vague. Subsequently, there have been a few techniques proposed that use Pareto domination based acceptance criterion in MOSA (Smith, 2006; Bandyopadhyay et al, 2008; Cao et al, 2016), the merit of which is that the domination status of the point is considered with respect to not only the current solution but also the archive of non-dominated solutions found so far. It has been widely demonstrated that simulated annealing algorithms are capable of finding multiple Pareto-optimal solutions in a single run.

*2.3. Reinforcement Learning Hyper-Heuristic*

The reinforcement learning hyper-heuristic strategy proposed in this paper can be divided into two parts, namely, heuristic selection and credit assignment. The goal is to design online strategies that are capable of autonomously selecting between different heuristics based on their credits (Burke et al, 2013).

Credit assignment rewards the heuristics online based on certain criterion, and the credits are thereafter fed to the heuristic selection strategy. It is similar to the reward assignment in reinforcement learning where the agent receives a numerical reward based on the success of an action's outcome. In this study, we develop a new credit assignment strategy based on hypervolume (Zitzler and Thiele, 1999) increments and the number of solutions newly generated to calculate the credit $c_{i,t}$,

$$c_{i,t} = e^{\frac{i(t)}{iter}} \cdot \left( \frac{\frac{HV(PF_t) - HV(PF_{t-1})}{HV(PF_{true})} + \frac{|PF_t| - |\bigcap(PF_t, PF_{t-1})|}{|PF_t|}}{i(t) - i(t-1)} \right) \quad (4)$$

where *iter* denotes the total number of iterations, $i(t)$ is the number of iterations that has been performed at epoch $t$ (i.e., the $t$th time heuristic selection has been conducted), $PF_t$ is the Pareto front at $t$, and $HV(*)$ approximates the hypervolume of the Pareto front in percentage using Monte Carlo approach through $N$ uniformly distributed samples within the bounded hyper-cuboid to alleviate the computational burden. Specifically,

$$HV(PF, r^*) = volume(\bigcup_{x \in PF} v(x, r^*)) \quad (5)$$

where $r^*$ is the reference point which is set to be 1.1 times the upper bound of the Pareto front in the *HV* calculation following the recommendations in literature (Ishibuchi et al., 2010; Li et al., 2016). Therefore, in Equation (4), $HV(PF_t) \in [0, 1]$ is the hypervolume of the Pareto front at $t$, $(HV(PF_t) - HV(PF_{t-1}))$ is the hypervolume increment since the last time the heuristics are selected, and $HV(PF_{true})$ is the normalization term. The term $\frac{|PF_t| - |\bigcap(PF_t, PF_{t-1})|}{|PF_t|} \in [0, 1]$ computes the percentage of newly generated solution in the current Pareto front. Both terms are dimensionless and they are summed together first then divided by $(i(t) - i(t-1))$ to evaluate the performance of a heuristic as reflected by the evolution of the Pareto front per iteration. Because it is easier for the optimizer to achieve improvements at early stage of optimization, we introduce the compensatory factor $e^{\frac{iter(t)}{iter}} \in [1, e]$ to progressively emphasize the credits earned as the optimization progresses.

Heuristic selection, as its name indicates, selects from the low-level heuristics at each time epoch. The concept is similar to agent in reinforcement learning. One difficulty that a heuristic selection strategy would have to overcome is the exploration versus exploitation dilemma (EvE), indicating that the heuristic with the highest credits should be favored whilst the heuristics with low credits should also be occasionally selected because they may produce high quality results as the search progresses. Many heuristic selection strategies have been proposed in literature, including probability matching (PM), adaptive pursuit (Thierens, 2007), choice function (Cowling et al, 2000; Maashi et al, 2015), Markov chain models (McClymont and Keedwell, 2011) and multi-armed bandit algorithms (Krempser et al, 2012). In this paper, we devise a heuristic selection strategy with minimal number of parameters inspired by the idea of probability matching to specifically fit the online learning scheme. Given a finite set of heuristic $\mathbf{O}$, an heuristic $o_i \in \mathbf{O}$ is selected at time $t$ with probability $p_{i,t}$ proportional to the heuristic's quality $q_{i,t}$, which is mainly determined by the credit $c_{i,t}$. The parameter $t$ is independent of the algorithm, indicating how many times the heuristic selection has been conducted. The update rule is given as follows,

$$q_{i,t} = \alpha \cdot q_{i,t-1} + (1-\alpha) \cdot c_{i,t} \quad (6)$$

$$p_{i,t} = p_{\min} + (1 - |\mathbf{O}| \cdot p_{\min}) \frac{q_{i,t}}{\sum_{j=1}^{|\mathbf{O}|} q_{j,t}} \quad (7)$$

where $p_{\min} \in (0, \frac{1}{|\mathbf{O}|}]$ is the minimum selection probability to facilitate exploration and guarantee $p_{i,t} \in [0, 1]$. It is greater than 0 so the heuristics with low credits are also considered. The forgetting factor $\alpha \in [0, 1]$ determines the importance of the credits received previously because the current solution may be the result of a decision taken in the past. In this study, $p_{\min}$ is chosen to be 0.1 and $\alpha$ is chosen to be 0.5. It is worth noting here again that $t-1$ in Equation (5) does not imply the iteration before $t$ in optimization; it means the last time the hyper-heuristic is updated. And we only update the values that correspond to the chosen heuristic at $t-1$. For unselected heuristics we have $q_{i,t} = q_{i,t-1}$. After $p_{i,t}$ is determined using Equations (5) and (6), the lower level heuristic is chosen per its probability using roulette wheel selection method.

## 3. HYPER-HEURISTIC MOSA

With the hyper-heuristic rules defined, the MOSA algorithm and the joint hyper-heuristic scheme are presented in this section.

*3.1. MOSA/R Algorithm*

The algorithm used in this study is referred to as MOSA/R (Multi-Objective Simulated Annealing based on Re-seed) which was originally developed and applied to configuration optimization (Cao et al, 2016). MOSA/R uses the concept of the amount of domination in computing the acceptance probability of a new solution. The algorithm was designed aiming at solving multi-modal optimization problems with strong constraints. It is capable of providing feasible solutions more efficiently compared to traditional MOSAs due to the re-seed technique developed. As will be demonstrated in this paper, the advancement of MOSA/R can be generalized with hyper-heuristic by making the re-seed step autonomously to cater towards various design preferences. The pseudo-code of MOSA/R is provided below.

---
**Algorithm MOSA/R**

Set *Tmax*, *Tmin*, # of iterations per temperature *iter*, cooling rate *α*, *k* = 0
Initialize the *Archive* (Pareto front)
*Current solution* = randomly chosen from *Archive*
**While** (*T* > *Tmin*)
    **For** 1 : *iter*
        Generate a *new solution* in the neighborhood of *current solution*
        **If** *new solution* dominates *k* (*k* >= 1) solutions in the *Archive*
            **Update**
        **Else if** *new solution* dominated by *k* solutions in the *Archive*
            **Action**
        **Else if** *new solution* non-dominant to *Archive*
            **Action**
        **End if**
    **End for**
    *k* = *k*+1
    *T* = (*α*$^k$)**Tmax*
**End While**

---
**Algorithm Update**

Remove all *k* dominated solutions from the *Archive*
Add *new solution* to the *Archive*
Set *new solution* as *current solution*

---
**Algorithm Action**

**If** *new solution* and *Archive* are non-dominant to each other
    Set *new solution* as *current solution*
**Else**
    **If** *new solution* dominated by *current solution*
      **Re-seed**
    **Else**
      **Simulated Annealing**
    **End If**
**End if**

---
**Algorithm Re-seed**

*new solution* is dominated by *k* (*k* >= 1) solutions in the *Archive*
Select a heuristic from **low-level heuristics** based on hyper-heuristic strategy
Set *selected solution* following the selected **heuristic**
**If** $\frac{1}{1+\exp(-\Delta dom_{selected,new}/\max(T,1))} > \text{rand}(0,1)$*
    Set *selected solution* as *current solution*
**Else**
    **Simulated Annealing**
**End if**
* rand(0,1) generates a random number between 0 to1

---
**Algorithm Simulated Annealing**

$\Delta dom_{avg} = \frac{\sum_{i=1}^{k}\Delta dom_{i,new}}{k}$

**If** $\frac{1}{1+\exp(\Delta dom_{avg}/T)} > \text{rand}(0,1)$
    Set *new solution* as *current solution*
**End if**

---

Given two solutions **a** and **b**, if $\mathbf{a} \prec \mathbf{b}$ then the amount of domination is defined as,

$$\Delta dom_{\mathbf{a,b}} = \prod_{i=1, f_i(\mathbf{a})\neq f_i(\mathbf{b})}^{M} (|f_i(\mathbf{a})-f_i(\mathbf{b})|/R_i) \qquad (8)$$

where *M* is the number of objectives and $R_i$ is the range of the *i*th objective (Bandyopadhyay et al, 2008). As indicated in the pseudo-code, the hyper-heuristic scheme comes into effect in Algorithm Re-seed. Whenever re-seed is triggered, first a low-level heuristic is selected from the repository based on the proposed reinforcement learning hyper-heuristic (Section 2.3), and then the current solution is altered using the selected low-level heuristic. Most simulated annealing related hyper-heuristic studies (Antunes et al, 2011; Bai et al, 2012; Burke et al, 2013;) use simulated annealing as the high-level heuristic to select from lower level heuristic repository to exploit multiple neighborhoods which can be regarded as variable neighborhood search mechanism. However, in this research, the proposed approach uses probability matching (PM) as the high-level heuristic and part of the MOSA/R as lower level heuristics which can be regarded as adaptive operator selection (Maturana et al, 2009). In the next sub-section, we propose four low-level heuristics for the hyper-heuristic MOSA.

*3.2. Low-Level Heuristics*

Hereafter we refer to the MOSA/R with hyper heuristic scheme as MOSA/R-HH. The hyper-heuristic scheme intervenes in the re-seed scheme (Algorithm Re-seed) which essentially differs MOSA/R-HH (hyper-heuristic MOSA/R) from other MOSA algorithms. In this paper we propose four re-seed strategies as low-level heuristics.

(1) The solution in the *Archive* with the minimum amount of domination. The first strategy selects the solution from *Archive* that corresponds to the minimum difference of domination amount with respect to the *new solution*. For $\forall \mathbf{x} \in Archive$ that dominates the *new solution*,

$$\mathbf{x}_{select} = \arg\min_{\mathbf{x}}(\Delta dom_{\mathbf{x},\mathbf{x}_{new}})$$
$$= \arg\min_{\mathbf{x}}\left(\prod_{i=1, f_i(\mathbf{x}) \neq f_i(\mathbf{x}_{new})}^{M} (|f_i(\mathbf{x}) - f_i(\mathbf{x}_{new})|/R_i)\right) \quad (9)$$

Then the selected solution is set as *current solution* with probability $\frac{1}{1+\exp(-\Delta dom_{selected,new}/\max(T,1))}$. The solution corresponding to the minimum difference of domination amount is chosen to avoid premature convergence. An example is given in Figure 1(a), the solution selected using this strategy corresponds the one in the *Archive* that dominates the current solution the least.

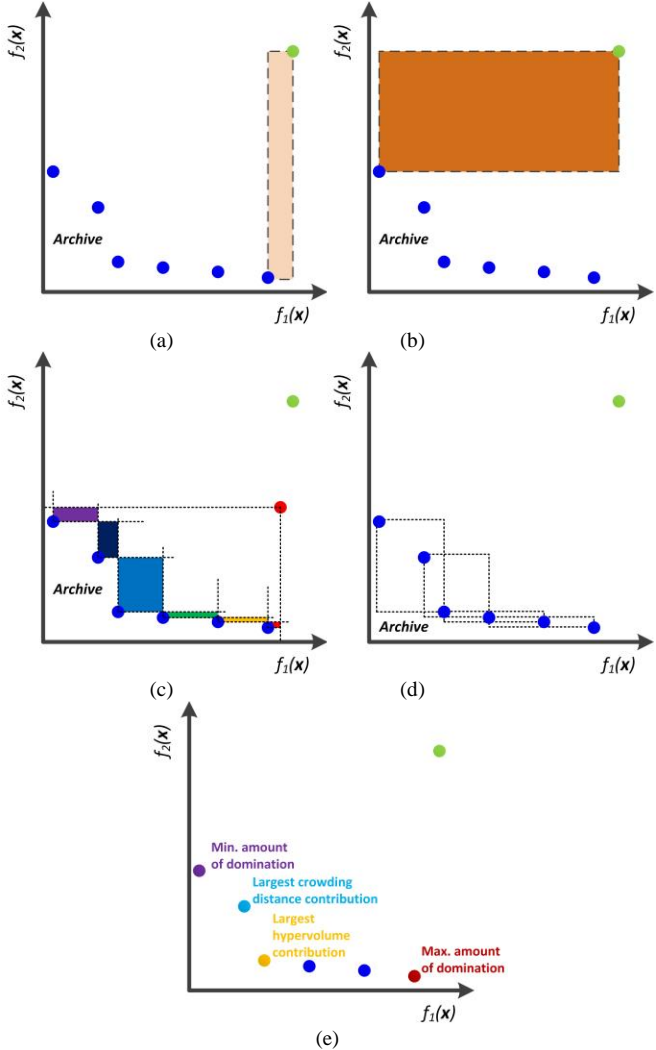

(a)  (b)

(c)  (d)

(e)
Figure 1 Examples of solutions selected by the four low-level heuristics

(2) The solution in the *Archive* with the maximum amount of domination. The second strategy is defined similarly to 1). For $\forall \mathbf{x} \in Archive$ that dominates the *new solution*,

$$\mathbf{x}_{select} = \arg\max_{\mathbf{x}}(\Delta dom_{\mathbf{x},\mathbf{x}_{new}})$$
$$= \arg\max_{\mathbf{x}}\left(\prod_{i=1, f_i(\mathbf{x}) \neq f_i(\mathbf{x}_{new})}^{M} (|f_i(\mathbf{x}) - f_i(\mathbf{x}_{new})|/R_i)\right) \quad (10)$$

The only difference is that this time the solution of the maximum domination amount compared to the *new solution* will be chosen. The strategy emphasizes more on the exploitation of better neighboring solutions compared to strategy (1) as (1) aims to maintain a balance between exploration and exploitation. The first two strategies are *new solution* dependent. Next we will introduce two new solution independent strategies. As shown in Figure 1(b), the selected solution using the second strategy dominates *current solution* the most.

(3) The solution in the *Archive* with the largest hypervolume (HV) contribution. In (3) we compute the hypervolume contribution of each point in *Archive* using the method proposed by Emmerich et al (2005). Hypervolume contribution quantifies how much each point in the Pareto front contributes to the HV as explained in Figure 1(c); the areas of the colored rectangles indicate the hypervolume contribution for each solution in the *Archive*. A large value of HV contribution indicates that the point stays in a less explored portion of the Pareto front whilst maintaining good convergent performance.

(4) The solution in the *Archive* with the largest crowding distance. The last strategy makes use of the technique called crowding distance (Deb et al, 2002). The point with the largest crowding distance will be selected. The strategy is inclined to exploration (diversity) in the EvE dilemma. As can be seen in Figure 1(d), in the minimization case, the crowding distance for each solution in the *Archive* is determined by the area of the bounding box formed by its adjacent solutions.

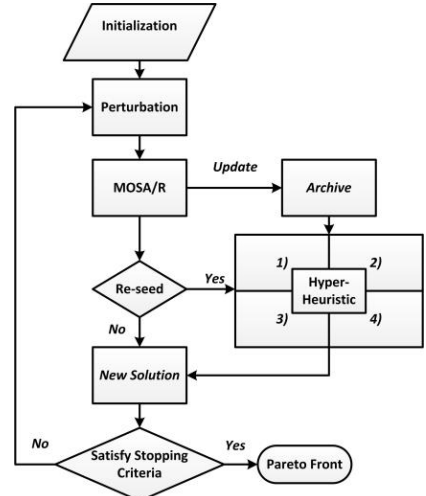

Figure 2 Flowchart of MOSA/R and embedded hyper-heuristic

As illustrated in Figure 1(e) which gives a comparison of the solutions selected by the proposed four low-level heuristics, each low-level heuristic designed has its own emphasis and intention. The hyper-heuristic scheme is designed to adaptively switch between different priorities that suits current search endeavor the best, and therefore could be applied to tackle different instances without further modification.

Figure 2 depicts the overall mechanism of MOSA/R and the co-acting hyper-heuristic in a flowchart.

## 4. BENCHMARK CASE STUDIES

*4.1. Test cases*

Here we apply four algorithms, the proposed algorithm MOSA/R-HH and three popular algorithms including an advanced multi-objective simulated annealing algorithm AMOSA (Bandyopadhyay et al, 2008), a fast and elitist multi-objective genetic algorithm NSGA-II (Deb et al, 2002), and A multi-objective evolutionary algorithm based on decomposition MOEA/D (Zhang and Li, 2007), to 14 benchmark test problems from DTLZ (Deb et al, 2002) and UF (Zhang et al., 2008) test suites. The three algorithms adopted for comparison are among the most recognized multi-objective algorithms and have been applied for a variety of optimization problems. The test sets are considered to be representative due to their diverse properties as listed in Table 2. All algorithms will be executed 5 times independently for each test problem.

Table 2 Benchmark test problem properties

| Instance | # Obj. | # Var. | Properties |
|---|---|---|---|
| DTLZ1 | 3 | 6 | Linear Pareto, multimodal |
| DTLZ2 | 3 | 7 | Concave Pareto |
| DTLZ3 | 3 | 10 | Concave Pareto, multimodal |
| DTLZ4 | 3 | 10 | Concave Pareto, biased solutions distribution |
| DTLZ5 | 3 | 10 | Concave degenerated Pareto |
| DTLZ6 | 3 | 10 | Concave Pareto, biased solutions distribution |
| DTLZ7 | 3 | 10 | Discontinuous Pareto |
| UF1 | 2 | 10 | Convex Pareto |
| UF2 | 2 | 10 | Convex Pareto |
| UF3 | 2 | 10 | Convex Pareto |
| UF4 | 2 | 10 | Concave Pareto |
| UF5 | 2 | 10 | Discrete Pareto |
| UF6 | 2 | 10 | Discontinuous Pareto, |
| UF7 | 2 | 10 | Linear Pareto |

*4.2. Parametric Setting*

The initial temperature is determined that virtually all solutions are accepted at the beginning 'burn in' period (Suman and Kumar, 2006). The stopping criterion, i.e., the final temperature, is chosen to control the error. In this research, the starting temperature $T_{max}$ and final temperature $T_{min}$ values of AMOSA and MOSA/R-HH are set to be 100 and $10^{-5}$, respectively. The total number of iterations, denoted as *iter*, is chosen to be 20,000 for DTLZ1 and DTLZ2, 30,000 for DTLZ3-7, and 100,000 for UF test instances. For temperature decrement $T = \Phi(T)$, we adopt the exponential approach,

$$T_{i+1} = \alpha^i T_i \qquad (11)$$

where $0 < \alpha < 1$ is chosen to be 0.8. Note that each parameter in AMOSA is set to be the same as that of MOSA/R-HH. For NSGA-II and MOEA/D, the total number of function evaluations is set in accordance with AMOSA and MOSA/R-HH. Other parameters used follow those used in literature (Deb et al, 2002; Zhang and Li, 2007). For 2-objective test problems, the population size is set to be 150, and 300 for 3-objective test problems. The distribution indices of Simulated Crossover (SBX) and polynomial mutation are set to be 20. The crossover rate is 1.00 and the mutation ration is $1/n$ where $n$ is the length of decision vector. In MOEA/D, Tchebycheff approach is used and the size of neighbor population is set to be 20. All initial solutions are generated randomly form the decision space of the problems.

*4.3. Performance Metrics*

For multi-objective optimization (MOO), an algorithm should provide a set of solutions that realize the optimal trade-offs between the considered optimization objectives, i.e., Pareto set. Therefore, the performance comparison of MOO algorithms is based on their Pareto sets. In this study, two popular metrics IGD and HV are used to quantify the performance of the algorithms.

*Inverted Generational Distance* (IGD)

The IGD indicator measures the degree of convergence by computing the average of the minimum distance of points in the true Pareto front (*PF\**) to points in Pareto front obtained (*PF*), as described below,

$$IGD(PF, PF^*) = \frac{\sum_{\mathbf{f}^* \in PF^*, i=1}^{|PF^*|} \sqrt{\min_{\mathbf{f} \in PF}\left(\sum_{m=1}^{M}(f_m^i * - f_m)^2\right)}}{|PF^*|} \qquad (12)$$

where *M* is the number of objectives, $f_m$ is the *m*-th objective value of $\mathbf{f} \in PF$. In Equation (12), $\min_{\mathbf{f} \in PF}\left(\sum_{m=1}^{M}(f_m^i * - f_m)^2\right)$ calculates the minimum Euclidean distance between the *i*th point in *PF\** and points in *PF*. A lower value of IGD indicates better convergence and completeness of the *PF* obtained.

*Hypervolume* (HV)

Refer to Equation (5). HV indicator measures convergence as well as diversity. The calculation of HV requires normalized objective function values and in this paper HV stands for the percentage covered by the Pareto front of the cuboid defined by the reference point and the original point (0, 0, 0). As mentioned earlier, the reference point is set to be 1.1 times the upper bound of the *PF\**.

*4.4. Test Case Results and Discussions*

The benchmark experiment examines the performance of MOSA/R-HH, AMOSA, NSGA-II, and MOEA/D as applied to DTLZ and UF test suites. The analysis results are based on 5 independent test runs. The mean and standard deviation of IGD and HV are recorded. All computations are carried out within MATLAB on a 2.40GHz Xeon E5620 desktop.

Tables 3 and 4 illustrate the relative performance of all four algorithms in terms of the two metrics IGD and HV where we keep 4 significant digits for mean and 3 significant figures for standard deviation. The shaded grids indicate the best result in each test in terms of the mean value. The performance comparison as well as the robustness of each algorithm are also illustrated in Figures 4. As can be observed from the figure, MOSA/R-HH prevails in DTLZ1, DTLZ2, DTLZ5 and DTLZ7 in both metrics. MOEA/D has an edge over MOSA/R-HH in DTLZ3, while MOSA/R-HH performs significantly better than NSGA-II and AMOSA. DTLZ4 is a close race for MOSA/R-HH, NSGA-II and MOEA/D. And for DTLZ6, MOSA/R-HH, AMOSA and MOEA/D all demonstrate similar performance. Figure 5 depicts the Pareto surface obtained by each algorithm when applied to DTLZ1 test case. For UF test cases, MOSA/R-HH takes the lead in three of them in both IGD and HV, which is the best among the four algorithms. Figure 6 shows an example of the Pareto front obtained by each algorithm for UF4 in comparison with the true Pareto front. It can be noticed that the Pareto front obtained by MOSA/R-HH stays close to the true Pareto front and maintains good diversity. The performance of AMOSA, NAGA-II and MOEA/D fluctuate as test instance changes due to different problem properties. On the other hand, MOSA/R-HH is more robust and outperforms other algorithms when tackling most test instances because of the adaptive hyper-heuristic scheme.

Table 3 Numerical test results: IGD mean and standard deviation

| Instance | MOSA/R-HH | AMOSA | NSGA-II | MOEA/D |
|---|---|---|---|---|
| DTLZ1 | 0.007191 (3.69E-4) | 0.02134 (0.00506) | 1.656 (0.538) | 0.01315 (0.00195) |
| DTLZ2 | 0.01403 (0.00127) | 0.01992 (0.00107) | 0.03093 (0.00147) | 0.02434 (0.00173) |
| DTLZ3 | 0.06330 (0.00380) | 0.7198 (0.131) | 7.419 (1.87) | 0.0342 (0.0125) |
| DTLZ4 | 0.02263 (0.00222) | 0.07643 (0.00456) | 0.02176 (0.000668) | 0.02334 (0.00176) |
| DTLZ5 | 6.356E-4 (4.34E-5) | 0.001956 (1.49E-4) | 0.001390 (2.74E-4) | 0.002541 (0.0966) |
| DTLZ6 | 3.231 E-4 (5.42E-6) | 4.404E-4 (1.85E-4) | 0.8738 (0.0762) | 0.001792 (2.20E-4) |
| DTLZ7 | 0.01657 (9.49E-4) | 0.01928 (5.45E-4) | 0.8235 (0.0211) | 0.06502 (0.00152) |
| UF1 | 0.01252 (0.00189) | 0.03509 (0.00250) | 0.01972 (0.00967) | 0.01938 (0.00567) |
| UF2 | 0.002974 (6.25E-4) | 0.005458 (8.87E-05) | 0.006871 (0.00365) | 0.01876 (0.00563) |
| UF3 | 0.2477 (0.104) | 0.3797 (0.368) | 0.1559 (0.0131) | 0.2553 (0.0323) |
| UF4 | 0.01905 (8.76E-4) | 0.03124 (1.99E-4) | 0.03792 (0.00397) | 0.04796 (0.00513) |
| UF5 | 0.1636 (0.00666) | 0.1523 (0.0242) | 0.6759 (0.279) | 0.6501 (0.292) |
| UF6 | 0.1412 (0.0816) | 0.09371 (4.34E-06) | 0.4929 (0.0963) | 0.5606 (0.151) |
| UF7 | 0.01713 (1.33 E-4) | 0.03393 (0.00514) | 0.008407 (0.00309) | 0.005269 (5.043E-4) |

Table 4 Numerical test results: HV mean and standard deviation

| Instance | MOSA/R-HH | AMOSA | NSGA-II | MOEA/D |
|---|---|---|---|---|
| DTLZ1 | 0.8593 (0.0204) | 0.8312 (0.0184) | 0.04210 (0.0941) | 0.8353 (0.0282) |
| DTLZ2 | 0.5945 (0.00586) | 0.5850 (0.00130) | 0.5663 (0.00832) | 0.5789 (0.00420) |
| DTLZ3 | 0.5280 (0.0380) | 0.004466 (0.00470) | 0.001404 (0.00236) | 0.5376 (0.0248) |
| DTLZ4 | 0.5739 (0.00869) | 0.5535 (0.00738) | 0.5686 (0.00765) | 0.5763 (0.00877) |
| DTLZ5 | 0.2139 (0.00157) | 0.2096 (0.00125) | 0.2097 (0.00100) | 0.2038 (0.00356) |
| DTLZ6 | 0.2059 (0.00568) | 0.2029 (0.00166) | 0.001440 (0.00211) | 0.2012 (0.00119) |
| DTLZ7 | 0.2635 (0.00549) | 0.2580 (0.0122) | 0.1683 (0.00304) | 0.2498 (0.00557) |
| UF1 | 0.7114 (0.00231) | 0.683 (0.00198) | 0.6958 (0.0126) | 0.6962 (6.37E-4) |
| UF2 | 0.7207 (5.52 E-4) | 0.71843 (4.03E-4) | 0.7165 (0.00351) | 0.7036 (0.00355) |
| UF3 | 0.4724 (0.0993) | 0.4098 (0.227) | 0.5196 (0.0204) | 0.3787 (0.0454) |
| UF4 | 0.4224 (0.00295) | 0.4044 (0.00659) | 0.3919 (0.00760) | 0.3885 (0.0131) |
| UF5 | 0.3613 (0.0346) | 0.3651 (0.0405) | 0.05647 0.0524 | 0.1128 (0.158) |
| UF6 | 0.3287 (0.0428) | 0.3487 (0.00766) | 0.1104 0.0413 | 0.2214 (0.0643) |
| UF7 | 0.5677 (0.00127) | 0.5454 (0.00541) | 0.5734 (0.00451) | 0.5773 (0.00169) |

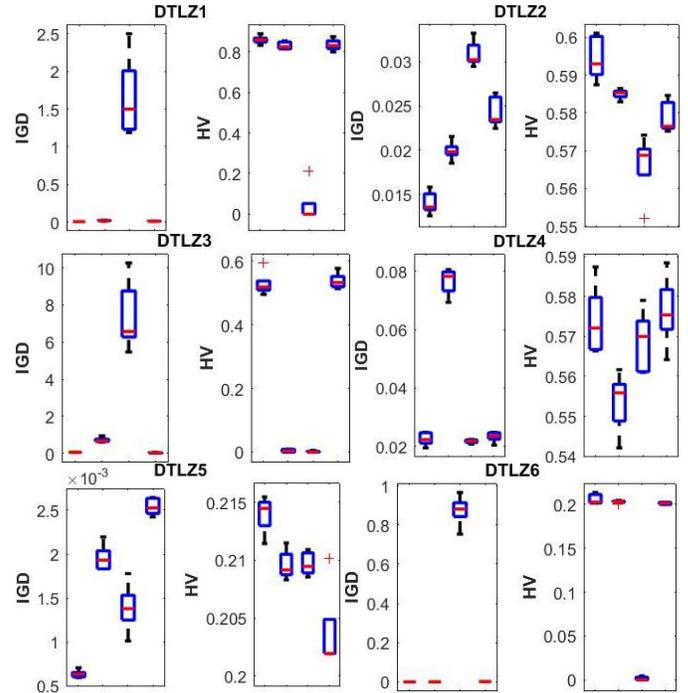

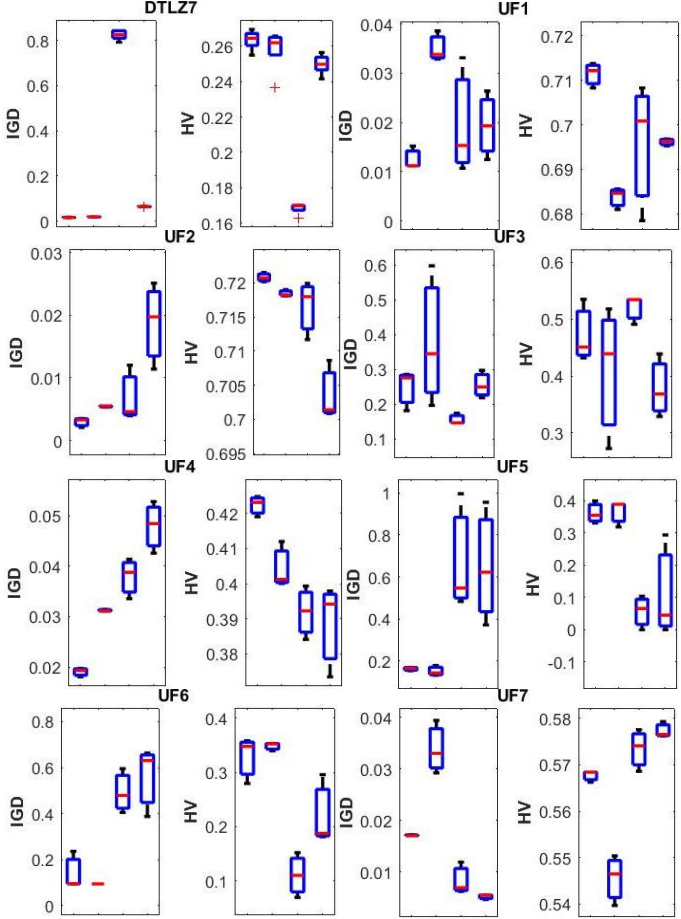

Figure 4 IGD and HV comparison of the four algorithms on the DTLZ and UF problems
(From left to right: MOSA/R-HH, AMOSA, NAGA-II, MOEA/D)

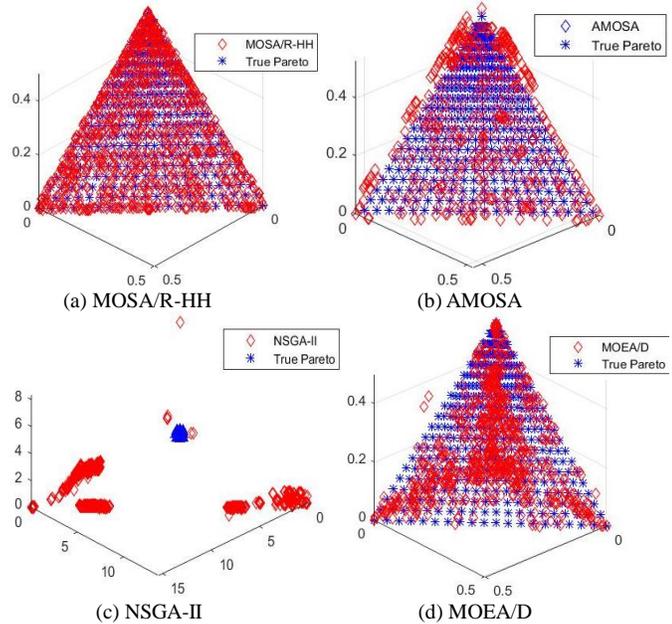

Figure 5 Pareto front obtained by each algorithm for test instance DTLZ1

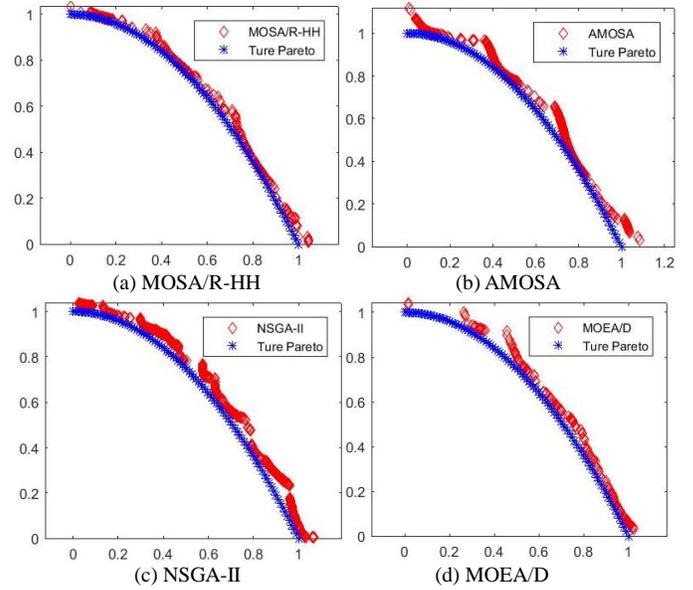

Figure 6 Pareto front obtained by each algorithm for test instance UF4

## 5. STRUCTURAL FAULT IDENTIFICATION USING MOSA/R-HH

In this section, we apply the proposed approach (MOSA/R-HH) and the original MOSA/R to a practical engineering problem, the identification of fault parameters in a structure, to showcase the advantage of incorporating the proposed hyper-heuristic technique. Structural fault identification is generally realized by inverse analysis through comparison between sensor measurements and model prediction in the parametric space. Here we specifically utilize the vibration response measurements (Cao et al, 2018a) because such identification naturally calls for a multi-objective optimization. We aim at solving the problem using the hyper-heuristic framework developed without taking advantage of any empirical domain-knowledge.

In model-based fault identification, a credible finite element model of the structure being monitored is available. The stiffness matrix of the structure under the healthy condition is denoted as $\mathbf{K}^R = \sum_{i=1}^{n} \mathbf{K}_i^R$, where $n$ is the number of elements, and $\mathbf{K}_i^R$ is the reference (healthy) stiffness of the $i$-th element. Without loss of generality, we assume that damage causes stiffness change. The stiffness matrix of the structure with fault is denoted as $\mathbf{K}^D = \sum_{i=1}^{n} \mathbf{K}_i^D$, where $\mathbf{K}_i^D = (1-\alpha_i)\mathbf{K}_i^R$. $\alpha_i \in [0, 1]$ ($i = 1, \cdots, n$) is the fault index for the $i$-th element. For example, if the $i$-th element suffers from damage that leads to a 20% of stiffness loss, then $\alpha_i = 0.2$. We further assume that the structure is lightly damped. The $j$-th eigenvalue (square of natural frequency) and the $j$-th mode (eigenvector) are related as $\lambda_j = \{\phi_j\}^T \mathbf{K} \{\phi_j\}$. The change of the $j$-th eigenvalue from the healthy status to the damaged status can be derived as (Cao et al, 2018a),

$$\Delta\lambda_j = \{\phi_j\}^T (\mathbf{K}^D - \mathbf{K}^R)\{\phi_j\} = \sum_{i=1}^{n} \alpha_i \{\phi_j\}^T \mathbf{K}_i^R \{\phi_j\} \quad (13)$$

which can be re-written as

$$\Delta\lambda_j = \sum_{i=1}^{n} \alpha_i \cdot S \quad (14)$$

or, in matrix/vector form,

$$\Delta\lambda = \mathbf{S}\alpha \quad (15)$$

where $\mathbf{S}$ is the sensitivity matrix whose elements are given in Equation (13), and $\Delta\lambda$ and $\alpha$ are, respectively, the $q$-dimensional natural frequency change vector (based on the comparison of measurements and baseline healthy results) and the $n$-dimensional fault index vector.

It is worth noting that the inverse identification problem (Equation (15)) is usually underdetermined in engineering practices, because $n$, the number of unknowns (i.e., the number of finite elements), is usually much greater than $q$, the number of natural frequencies that can be realistically measured. This serves as the main reason that we want to avoid matrix inversion of $\mathbf{S}$ and resort to optimization by minimizing the difference between the measurements and predictions obtained from a model with sampled fault index values. In this study, we adopt a correlation coefficient, referred to as the multiple damage location assurance criterion (MDLAC) (Messina et al, 1998; Barthorpe et al, 2017; Cao et al, 2018a), to compare two natural frequency change vectors, as expressed below,

$$MDLAC(\Delta\lambda, \alpha) = \frac{\langle \Delta\lambda, \delta\lambda(\alpha) \rangle^2}{\langle \Delta\lambda, \Delta\lambda \rangle \cdot \langle \delta\lambda(\alpha), \delta\lambda(\alpha) \rangle} \quad (16)$$

where $\langle *, * \rangle$ calculates the inner product of two vectors. $MDLAC(\Delta\lambda, \alpha) \in [0,1]$ captures the similarity between measured frequency change $\Delta\lambda$ and predicted frequency change $\delta\lambda$. Furthermore, in addition to natural frequency change information, we also take into consideration the mode shape change information. For the $j$-th mode shape which itself is a vector, we can compare the measured change and predicted change using MDLAC in a similar manner. Therefore, a multi-objective minimization problem for an $n$-element structure can be formulated as following,

Find: $\alpha = \{\alpha_1, \alpha_2, ..., \alpha_n\}$

Minimize: $f_1 = -MDLAC(\Delta\lambda, \alpha)$,
$f_2 = -MDLAC(\Delta\phi, \alpha)$

Subject to: $\alpha^l \leq \alpha_i \leq \alpha^u$ (17)

where $\alpha^l$ and $\alpha^u$ are the pre-specified lower bound and upper bound of the fault index. The optimization problem defined above is non-convex. For notation simplicity, here without loss of generality we assume one mode is being measured and the information of mode shape change, denoted as $\Delta\phi$, is employed fault identification. In practical applications, multiple modes can be measured and compared.

Prior and empirical knowledge often plays an important role when tackling this type of structural fault identification problem due to infinitely many combinations of possible fault patterns. For example, some studies assume the number of faults is known beforehand (Shuai et al, 2017). Some other investigations take advantage of the sparse nature of the fault indices (Huang et al, 2017; Cao et al, 2018b). In this study, we apply the multi-objective simulated annealing to identify the fault pattern in terms of $\alpha$. We demonstrate the effectiveness of the adaptive hyper-heuristic approach, whereas we do not exploit any domain knowledge. To facilitate easy re-production of case analyses for interested readers, a benchmark cantilever beam model with varying number of elements and different fault patterns is used in this case demonstration. The Young's modulus of the beam is 69 GPa, the length per element is set as 10 m, and the area of cross-section is 1 $m^3$. The measurements of mode shape and natural frequencies used are simulated directly from the finite element models which are subject to 2‰ standard Gaussian uncertainties. Hereafter the measurements available to fault identification are limited to the first 5 natural frequencies and the $2^{nd}$ mode shape.

*5.1. Case Study 1: 20 elements, 2 faults*

We first carry out the case study on a 20-element cantilever beam. The faults are on the $6^{th}$ and $11^{th}$ element with severities $\alpha_6 = 0.04$ and $\alpha_{11} = 0.06$, respectively.

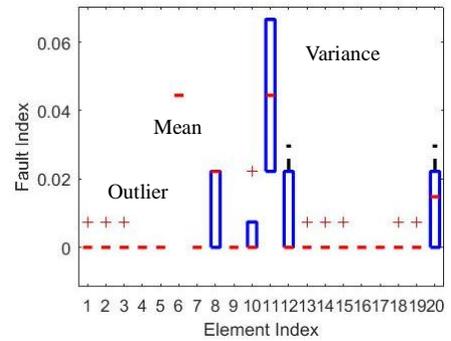
(a) MOSA/R: box plot

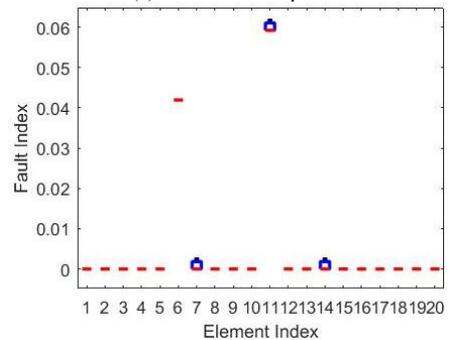
(b) MOSA/R-HH: box plot

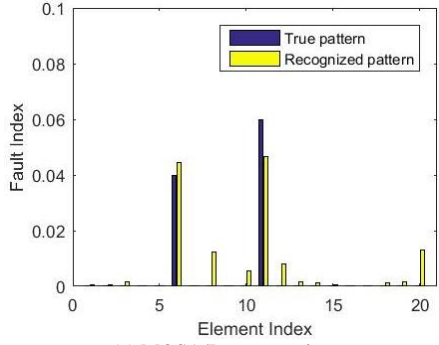
(c) MOSA/R: mean value

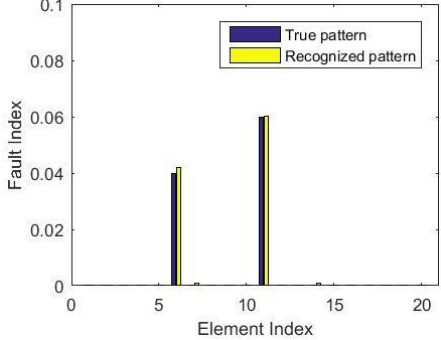
(d) MOSA/R-HH: mean value
Figure 7 Case study 1: fault identification results using MOSA/R and MOSA/R-HH

MOSA/R-HH proposed and MOSA/R are applied without knowing the number of faults. A set of optimal candidates are obtained, owing to the tradeoff between objectives. Each solution obtained corresponds to one possible fault pattern. Figure 7(a) and Figure 7(b) show the mean and variance of the solution sets with respect to the fault index for each element, in which mean value is represented by a dash, variance is depicted as a box, and plus sign stands for outlier. The uncertainty and fluctuation of the results mainly come from the noise introduced to the measurements and the under-determined nature of the problem. As seen in the figures, the results of MOSA/R-HH are more robust with fewer outliers. The mean values are then compared to the true fault pattern indices in Figure 7(c) and Figure 7(d). As illustrated, MOSA-HH is able to identify the location and severity of the fault pattern with better performance compared to MOSA/R due to the incorporated reinforcement learning hyper-heuristic. It adaptively adjusts the search direction as it progresses to yield a solution set of better distribution and accuracy.

### 5.2. Case Study 2: 30 elements, 3 faults

In the second case study, we perform a more difficult fault identification investigation using a 30-element cantilever beam. Three elements ($6^{th}$, $11^{th}$ and $22^{nd}$) are subject to faults with severities $\alpha_6 = 0.04$, $\alpha_{11} = 0.06$ and $\alpha_{22} = 0.02$ respectively. Compared to the case study conducted in Section 5.1, the case presented in this section is more challenging because of the many more possible combinations of fault patterns, the number of which grows exponentially with the number of elements.

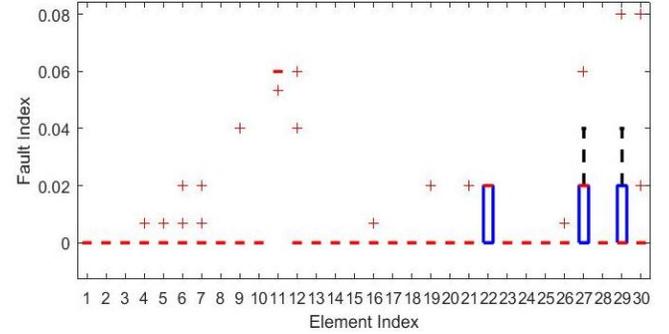
(a) MOSA/R: box plot

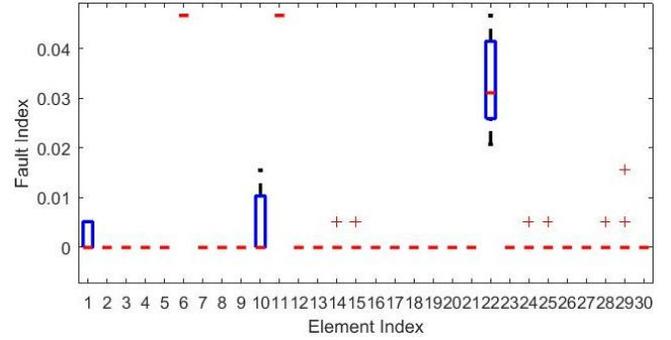
(b) MOSA/R-HH: box plot

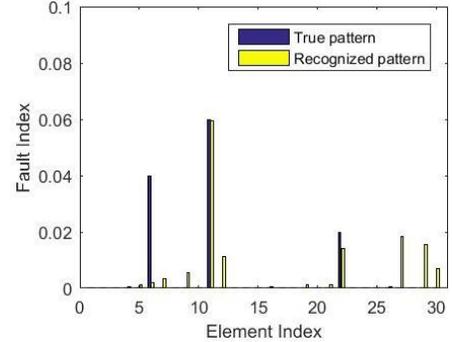
(c) MOSA/R: mean value

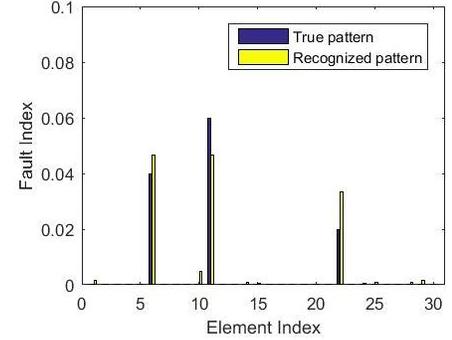
(d) MOSA/R-HH: mean value
Figure 8 Case study 2: fault identification results using MOSA/R and MOSA/R-HH

The proposed MOSA/R-HH is still capable of identifying a set of optimal solutions as possible fault patterns. Figure 8(a) and Figure 8(b) compare the mean and variance of the optimal solutions generated using MOSA/R and MOSA/R-HH. As observed, the result set of MOSA/R-HH is more consistent and thus has fewer outliers. Due to the enlarged search space, the solutions tend to have larger variance compared to that reported in Section 5.1. The mean value is then compared to the true fault pattern in Figure 8(c) and Figure 8(d). MOSA/R-HH demonstrates better performance compared to MOSA/R. For an ideal model without uncertainty, adding variables (number of elements in the structure) alone would not change the essence of the problem. In other words, if the optimization process lasts long enough, the quality of the final solutions would not deteriorate. However, errors and uncertainties are inevitable in engineering practices and play important role in our simulation. Nevertheless, MOSA/R-HH is capable of identifying the fault pattern in terms of both location and severity while MOSA/R, in this case investigation, completely overlooks the fault on the $6^{th}$ element (Figure 8(c)). The mean values of the MOSA/R-HH results bear some small errors but the overall fault pattern is practically recognized without using domain knowledge.

## 6. CONCLUDING REMARKS

In this research, we formulate an autonomous hyper-heuristic scheme that works coherently with multi-objective simulated annealing, featuring domination amount, crowding distance and hypervolume calculations. The hyper-heuristic scheme can be adjusted at high-level by changing heuristic selection and credit assignment strategies or at low-level by customizing the heuristic repository to meet different optimization requirements. It can also be used to investigate the relation between heuristics and problem instances. The proposed MOSA/R-HH yields better results than other MOSA algorithm like AMOSA and representative evolutionary algorithms like NSGA-II and MOEA/D when applied to benchmark test cases. Hyper-heuristic methodology is promising as it can address the problem adaptively based on defined low-level heuristics and on-line performance evaluation. The proposed hyper-heuristic approach is successfully devised to solve a representative structural fault identification problem without using any domain knowledge, as the hyper-heuristic framework autonomous adjusts the search iteratively during search. Due to the adaptive nature of the proposed methodology, the newly proposed framework can be extended to a variety of design and manufacturing optimization applications.

## ACKNOWLEDGMENT

This research is supported by in part by AFOSR under grant FA9550-14-1-0384 and in part by NSF under grant IIS-1741171.